# Zero-Shot Scene Understanding with Multimodal Large Language Models for Automated Vehicles

Mohammed Elhenawy
*CARRS-Q*
*Queensland University of Technology*
Brisbane, Australia
mohammed.elhenawy@qut.edu.au

Shadi Jaradat
*CARRS-Q*
*Queensland University of Technology*
Brisbane, Australia
shadi.jaradat@hdr.qut.edu.au

Taqwa I. Alhadidi
*Civil Engineering Department*
*Al-Ahliyya Amman University*
Amman, Jordan
t.alhadidi@ammanu.edu.jo

Huthaifa I. Ashqar
*Data Science and AI Department*
*Arab American University*
Ramallah, Palestine
huthaifa.ashqar@aaup.edu

Ahmed Jaber
Department of Transport Technology and Economics
*Budapest University of Technology and Economics*
Műegyetem Rkp. 3., H-1111 Budapest, Hungary
ahjaber6@edu.bme.hu

Andry Rakotonirainy
*CARRS-Q*
*Queensland University of Technology*
Brisbane, Australia
r.andry@qut.edu.au

Mohammad Abu Tami
*Natural, Engineering and Technology Sciences Department*
*Arab American University*
Ramallah, Palestine
mabutame@gmail.com

*Abstract*—Scene understanding is critical for various downstream tasks in autonomous driving, including facilitating driver-agent communication and enhancing human-centered explainability of autonomous vehicle (AV) decisions. This paper evaluates the capability of four multimodal large language models (MLLMs), including relatively small models, to understand scenes in a zero-shot, in-context learning setting. Additionally, we explore whether combining these models using an ensemble approach with majority voting can enhance scene understanding performance. Our experiments demonstrate that GPT-4o, the largest model, outperforms the others in scene understanding. However, the performance gap between GPT-4o and the smaller models is relatively modest, suggesting that advanced techniques such as improved in-context learning, retrieval-augmented generation (RAG), or fine-tuning could further optimize the smaller models' performance. We also observe mixed results with the ensemble approach: while some scene attributes show improvement in performance metrics such as F1-score, others experience a decline. These findings highlight the need for more sophisticated ensemble techniques to achieve consistent gains across all scene attributes. This study underscores the potential of leveraging MLLMs for scene understanding and provides insights into optimizing their performance for autonomous driving applications.

## I. INTRODUCTION

Vision-language models (VLM) are key components in autonomous driving and dynamic scene understanding. They incorporate visual input and language context into perception. Their integration allows a better interpretation of scenes [1], [2], [3]. Researchers showed that VLM can mimic human perception by recognizing the driving circumstances [4], VLM emphasizes on the importance of predicting driver focus for VLMs to use in emphasizing relevant visual data[5]. VLMs could mimic human perception by recognizing context [4]. This is crucial for safe navigation in dynamic environments with pedestrians, vehicles, and obstacles. The application of simultaneous localization and mapping (SLAM) further underscores the importance of integrating visual inputs with light detection and ranging (LiDAR) data for creating accurate environmental maps [6]. VLMs improve SLAM by providing contextual information to better interpret visual data, which improves localization and mapping, especially in urban environments. Recent advancements in deep learning have demonstrated the potential of VLMs to refine complex driving scene classifications [7]. Zou et al. emphasize the importance of robust lane detection methods for varying conditions[8] Furthermore, VLMs are instrumental in recognizing static entities such as pedestrians and vehicles, enhancing the understanding of human behavior in traffic [9]. VLMs enhance the classification of road elements, such as lane markings[8] and traffic signs [10], which are critical for safe navigation. Zou et al. emphasize robust lane detection methods for varying conditions [8].

This study aims to advance the application of MLLMs in Intelligent Transportation Systems (ITS), particularly for traffic scene understanding from an ego vehicle's perspective. A novel standardized framework for traffic scene analysis utilizing JSON-formatted prompts is introduced, optimizing frame-based analysis for real-time applications. This innovation establishes a foundation for expanding MLLM capabilities in traffic management. To contextualize these advancements, the following section examines the foundational technologies underpinning this research, including ADAS, AI-driven sensor fusion, and multimodal perception, which are critical for integrating MLLMs into autonomous driving systems.

## II. LITERATURE REVIEW

The integration of AI into Advanced Driver Assistance Systems (ADAS) has proven instrumental in monitoring

driver behavior and providing timely feedback or intervention to enhance safety [11]. By processing and analysing data from various sensors like cameras and LiDAR, AI substantially improves vehicular situational awareness and subsequent decision-making processes [12]. Additionally, understanding driver acceptance of ADAS can be enhanced by gaining insights into how drivers respond to automated systems, particularly in terms of reliance and trust [13], [14], [15]. Consequently, the design of ADAS must prioritize synergy between humans and machines by addressing drivers' mental models and interactions with the technology.

Driver training and knowledge of ADAS capabilities and limitations also play a critical role in shaping trust and acceptance. According to research, well-structured training programs can enhance drivers' perceived understanding of these systems, thereby improving their effectiveness in real-world applications [16] . As the reliance on ADAS continues to grow, it is crucial to consider how drivers interact with these systems, particularly in terms of their expectations and experiences. Effective ADAS design must account for not only the technological capabilities of the systems but also the mental models and cognitive load of the users [17]. These factors, alongside advancements in multimodal perception and explainability features, collectively contribute to the development of systems capable of responding to the dynamic nature of driving environments while ensuring both safety and efficiency.

Recent advancements in multimodal sensor fusion have further enhanced ADAS capabilities. Systems can achieve more robust and accurate environmental perception by integrating data from heterogeneous sensors such as cameras, LiDAR, and radar. For instance, Huang et al. provide a comprehensive survey on multimodal fusion techniques for autonomous driving perception, highlighting the challenges and solutions in combining data from different sensor modalities [18]. Moreover, developing condition-aware fusion methods allows ADAS to adapt to varying environmental conditions, thereby improving reliability and performance. The explainability of AI models within ADAS is another critical area of research. Ensuring that system decisions are transparent and understandable to users fosters trust and facilitates better human-machine interaction. Additionally, ensemble learning methods, such as those leveraging pre-trained transformers, have demonstrated potential for improving classification accuracy and robustness in complex scenarios, such as crash severity classification [19], [20]. These approaches are vital for building trust and ensuring that drivers can effectively engage with and understand automated systems [21], [22].

Zero-shot scene understanding using MLLMs has emerged as a promising approach for enhancing autonomous vehicle perception and decision-making capabilities. MLLMs, which combine the power of large language models with multimodal inputs, have shown impressive performance in various tasks, including scene understanding and object detection [23], [24]. In the context of autonomous vehicles, scene understanding plays a crucial role in enabling vehicles to perceive and interpret their surroundings accurately. Traditional approaches to scene understanding have relied on specialized computer vision techniques and sensor fusion methods. However, the integration of MLLMs offers new possibilities for more comprehensive and flexible scene interpretation [25], [26], [27].

One of the key advantages of using MLLMs for zero-shot scene understanding is their ability to generalize across different scenarios and environments without requiring extensive training on specific datasets [23], [24]. This is particularly valuable in the context of autonomous driving, where vehicles may encounter diverse and unpredictable situations. Recent studies have demonstrated the potential of MLLMs in various aspects of autonomous vehicle perception. For instance, researchers have explored the use of MLLMs for tasks such as object detection, semantic segmentation, and even OCR-free natural language processing of visual information. These capabilities can significantly enhance the situational awareness of autonomous vehicles, enabling them to better understand and respond to their environment [23], [24].

The integration of multiple sensor modalities, such as cameras, LiDAR, and radar, is crucial for robust scene understanding in autonomous vehicles. MLLMs have shown promise in effectively fusing information from these diverse sources, potentially leading to more accurate and reliable perception systems [23], [24]. This multimodal approach can help overcome the limitations of individual sensors and provide a more comprehensive understanding of the driving environment. Building on these insights, this study employs a structured methodology to evaluate the performance of MLLMs in zero-shot traffic scene understanding, leveraging ensemble techniques to address identified challenges and optimize model capabilities.

### III. METHODOLOGY

The section describes the research methodology along with the proposed framework in detail. This includes a structured evaluation process for analyzing the traffic scene and understanding the capability of MLLMs using the Honda Scenes Dataset (HSD). Fig.1. shows the methodology that will be used in a systematic way to increase robustness and Precision in traffic scene analysis by advanced AI techniques.

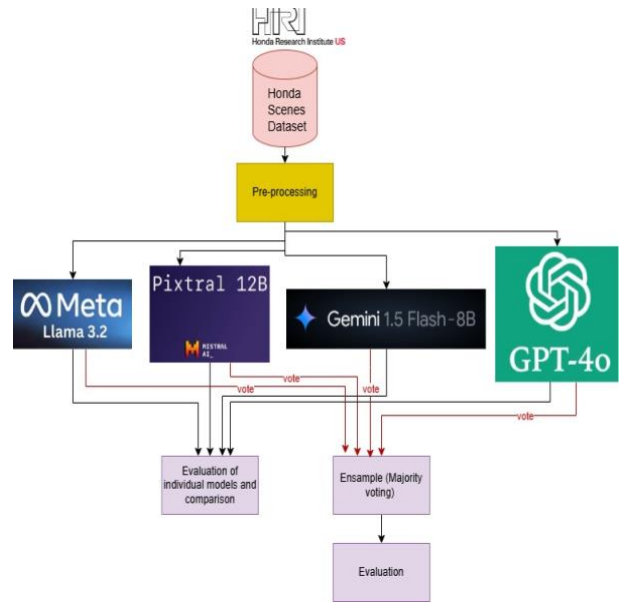

Fig.1: Evaluation workflow of MLLMs for traffic scene understanding and ensemble analysis.

The methodology involves several definite steps to ensure comprehensive assessment. These steps are listed below:

## A. Dataset Preparation and Preprocessing

First, the Honda Scenes Dataset was prepared with much care, for which high-quality driving videos were converted into frames, downscaling them to one frame per second to optimize processing. Frames with a lack of significant annotations or including stationary vehicles are excluded, hence arriving at a refined dataset of 5,006 frames, thereby prepared for detailed analysis.

## B. Model Selection

This analysis focuses on four modern, top MLLMs, which include Llama-3.2-11B-Vision, Pixtral 12B, Gemini 1.5 Flash-8B, and GPT-4o. Thus, this represents an ideal cross-section with significant varieties across model architecture and power; comprehensive analysis was expected through multiple testing.

## C. Appraisal Structure

Each model has to process the preprocessed dataset in a zero-shot, in-context learning format. This step leverages the inherent analytical capabilities of the MLLMs, which means they have been shown to interpret traffic scenes directly without any prior fine tuning; underlining once more models' adaptability and immediate applicative value.

## D. Ensemble Approach

An ensemble method using majority voting is applied to check for possible improvements in the performance. In this case, each model 'votes' according to its knowledge of particular scene attributes. This ensemble method seeks to combine diverse insights into a consensus that may raise the accuracy of the scene analysis.

## E. Evaluation and Comparison

The performance of each model will be evaluated on comprehensive metrics such as the F1-score, both individually and within the ensemble framework. This step is important in identifying the strengths and weaknesses of each model and in validating the effectiveness of the ensemble approach in improving overall scene understanding.

## F. Experimental Work and Prompt Engineering

The experimental setup further extends to prompt engineering, which involves designing specific multi-choice questions that systematically categorize the detected features from images captured from an ego vehicle's perspective. This provides a structured approach that not only standardizes the analysis but also enhances the granularity and specificity of the data extraction crucial for subsequent machine learning applications. The prompt is available in appendix A.

The structured prompt framework for processing scenes seen from an ego vehicle forms a strong tool in the analysis of driving-related images, explicitly to understand traffic scenes comprehensively. The wide range of environmental and road conditions falls within this framework by systematically categorizing detected features into 21 predefined attributes. It includes simple classifications of "Ambient"-describing light conditions to more complex scenarios: "Weather" conditions or highly elaborative road features with inclusion of intersections and zones of construction. This structured approach will ensure consistency and standardization occur during data analysis, and it makes the categorization process clear and applicable in every scenario. The output is in JSON format, and it will be much easier to integrate into a machine learning pipeline and compatible with an automated evaluation system. It is recommended because further analysis might easily be done using such a format and further manipulation and analyses are easy to manage within different AI-driven traffic and transportation research frameworks.

## IV. RESULTS

In this section, we test the performance of each model individually on the prompt from the previous section. But before going to the result analysis, we would like to note that our work is based on a frame-based approach. Since the prompt outlines attributes with temporal features-for example, "Stages" that include Approaching, Entering, Passing-such attributes will be simplified in a binary form for clarity and simplicity. For instance, categories such as "Rail Crossing" are represented as '0' in case of no detection and '1' upon detection. In effect, the temporal stages get consolidated into one binary outcome. This conversion aligns with the frame-based nature of our analysis, prioritizing the detection and classification of individual frames rather than temporal sequences. With this smooth representation, the framework has been optimized to work best in real-time and retrospective evaluations that make for effective and clear analysis of the attributes within a traffic scene.

The performance of four large language models on the attributes of a traffic scene presented in the radar charts has been quantified by the weighted average F1-Score, Recall, and Precision these results are shown in Fig.2. Overall, most of them perform similarly considering many attributes. However, the best performance of each one is in several different particular aspects; for example, GPT yields outstanding performance for complex attributes such as "Zebra Crossing," though Pixtral does the opposite in attributes like "Intersection (3-way)," showcasing its strength in scenarios requiring localised contextual understanding. Similarly, Gemini also shows consistent performances on standard features like "Surface" and "Stop Intersection," while the generally weaker performance of Llama turns in quite a decent performance with more straightforward attributes.

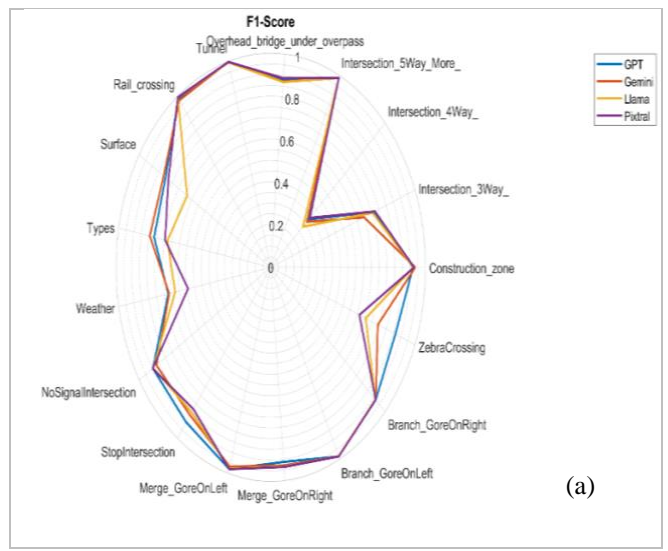

(a)

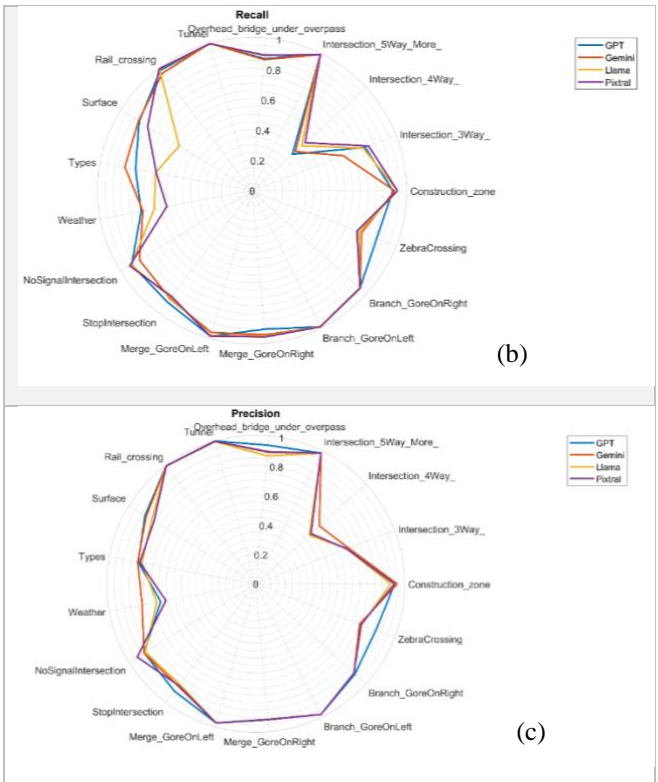

Fig.2: Models performance using (a) F1-score, (b) Recall, and (c) Precision.

These results reflect that some particular strengths different models have might be used with different tasks, indicating that specific fine-tuning for each model may be possible by observing these variations in performance. Thus, if prompt designing is done with regard to strengths of individual models, their performances for certain attributes can be further enhanced. It also underlines the flexibility that exists in improving model performance and compensating for its weaknesses in understanding scenes through prompt engineering. So, although overall performance may appear similar, individual model fine-tuning with prompt design may present a more feasible and balanced platform for traffic scene analysis. The average performance for each model in the different driving conditions is shown in Fig.3.

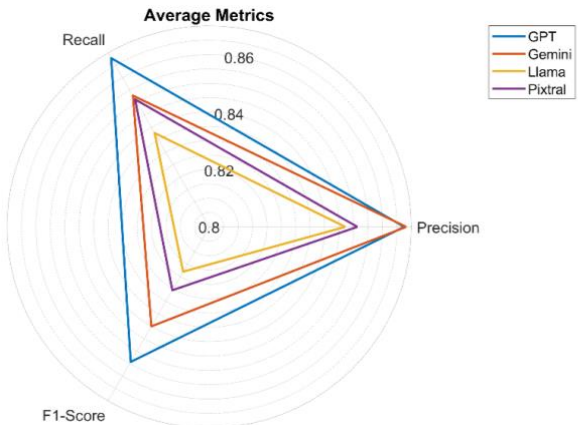

Fig. 3: Average Models performance

Fig. 3. shows the average metrics of GPT, Gemini, Llama, and Pixtral for the four multimodal large language models. The figure presents weighted averages of the metrics over all the traffic scene attributes evaluated and thus provides a high-level comparison of how each model performs, showing overall differences. This figure shows that GPT is the best in all measures, with an especially high Recall, which means it is better at catching relevant attributes. Pixtral and Gemini behave comparably, though Gemini has a slight edge over Pixtral on Precision and F1-score. As for Llama, although lagging behind, it still shows reasonable performance, especially on F1-Score, indicating that it can still keep a good balance between Recall and Precision.

This comparison done concisely helps to grasp the general strengths and weaknesses of each model. While GPT consistently leads in this, the spread across the different models is relatively modest, suggesting there is plenty of room for further optimization. Tailoring prompts or fine-tuning could help narrow the performance gaps for specific tasks or attributes. Overall, the figure provides a high-level summary of the models' capabilities in traffic scene analysis.

*Ensemble evaluation*

The analysis of individual models reveals that certain models outperform others on specific attributes, suggesting that employing a simple majority voting mechanism could enhance overall performance are sown in Fig.4. Consequently, we experimented with various combinations of three models as well as the full ensemble of all four models to evaluate the effectiveness of this approach. The figures illustrate the evaluation of ensemble approaches for weighted average Precision, Recall, and F1-Score differences compared to GPT, focusing on various traffic scene attributes. The ensembles combine outputs from different models using a simple majority voting approach to assess whether leveraging complementary model strengths can improve overall performance. The results show mixed outcomes, highlighting that while some attributes benefit from the ensemble approach, others experience performance declines. For instance, in Precision, attributes like "Intersection (3-way)" and "Weather" see improvements across multiple ensemble combinations, suggesting that majority voting effectively leverages the diverse strengths of individual models in these cases. Conversely, attributes such as "Zebra Crossing" and "Merge Gore on Left" exhibit reduced Precision, indicating that simple voting may not account for the complexities required to handle these scenarios effectively. Similarly, for Recall, improvements are observed for attributes like "Intersection (5-way or more)" and "Types," where the ensemble approach enhances the models' ability to capture relevant instances. However, for attributes like "Zebra Crossing," the Recall significantly drops, suggesting that misaligned outputs from individual models can negatively impact ensemble performance. The F1-Score results also reflect these trends. While ensemble combinations enhance balanced performance for attributes like "Weather" and "Surface," others, such as "Zebra Crossing," show substantial declines, indicating that majority voting struggles to handle nuanced attributes effectively. These findings suggest that while simple majority voting is effective for certain attributes, it is insufficient for consistently aggregating outputs across all attributes. This highlights the need for more advanced ensemble techniques, such as weighted voting, confidence-based aggregation, or fine-tuned combinations, to better handle the complexities of specific attributes and improve overall scene understanding performance.

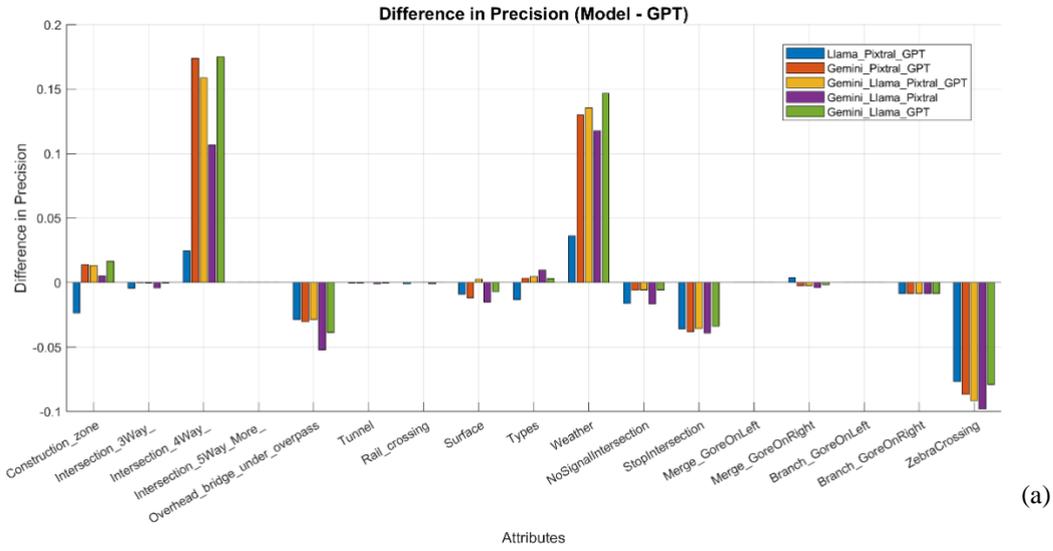

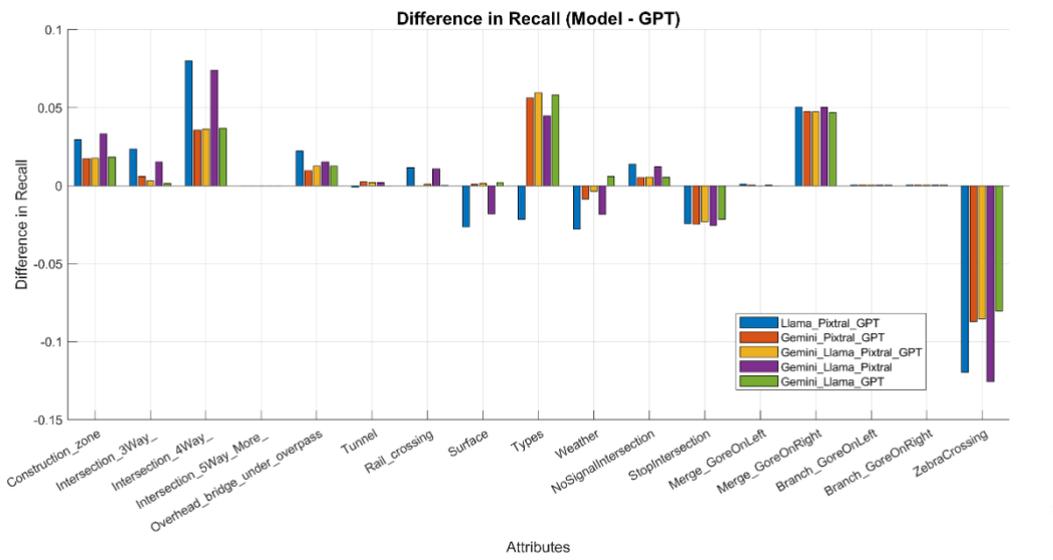

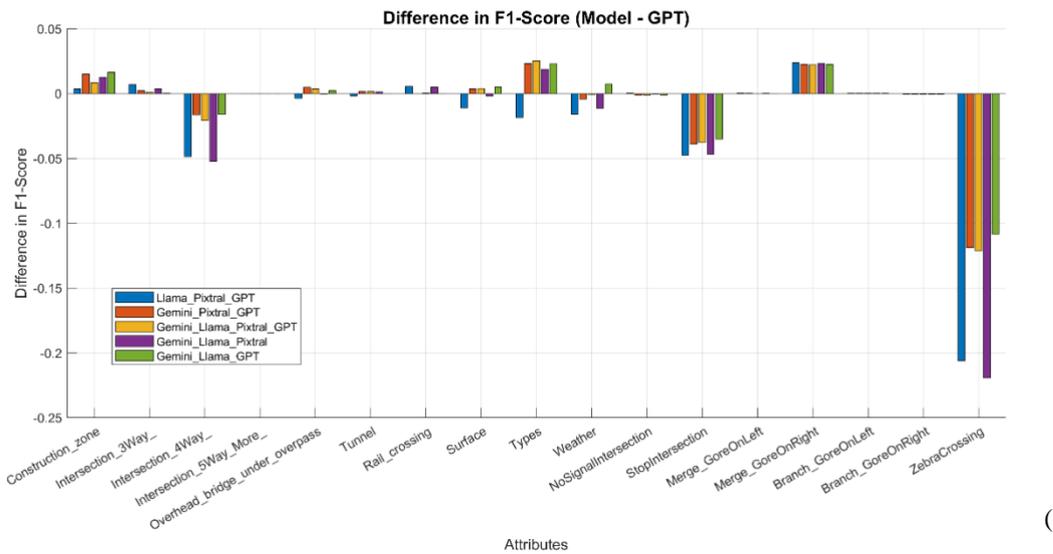

Fig. 4: Ensemble Models performance using (a) F1-score, (b) Recall, and (c) Precision.

## V. CONCLUSION

This work represents a massive leap forward in the research into traffic scene understanding using MLLMs. The structured evaluation framework in this study has shown that such models can indeed make sense of complex driving environments from the viewpoint of an ego vehicle. By systematically categorizing the detected features into predefined attributes such as road conditions, environmental factors, and complex road structures, it has been observed that MLLMs can go on to extract nuanced insights related to dynamic scenes in traffic.

These results from this study bring important insights into the capabilities and limitations of state-of-the-art AI technologies for traffic scene analysis and provide a stepping stone for further studies. Integrating such models with other AI technologies, expanding the category of attributes, or applying this framework within different driving contexts could be some possible future research that might improve the robustness and application of the traffic scene understanding systems.

ACKNOWLEDGMENT

This research was funded partially by the Australian Government through the Australian Research Council Discovery Project DP220102598.

REFERENCES

[1] Z. Zhong et al., "ClipCrop: Conditioned Cropping Driven by Vision-Language Model," in *Proceedings of the IEEE/CVF International Conference on Computer Vision*, 2023, pp. 294–304.

[2] X. Zhou, M. Liu, B. L. Zagar, E. Yurtsever, and A. C. Knoll, *Vision language models in autonomous driving and intelligent transportation systems*. 2023.

[3] M. A. Tami, H. I. Ashqar, M. Elhenawy, S. Glaser, and A. Rakotonirainy, "Using Multimodal Large Language Models (MLLMs) for Automated Detection of Traffic Safety-Critical Events," *Vehicles*, vol. 6, no. 3, pp. 1571–1590, 2024.

[4] C. Han et al., "Unbiased 3d semantic scene graph prediction in point cloud using deep learning," *Applied Sciences*, vol. 13, no. 9, p. 5657, 2023, doi: 10.3390/app13095657.

[5] A. Palazzi, D. Abati, S. Calderara, F. Solera, and R. Cucchiara, "Predicting the driver's focus of attention: The DR(eye)VE project," *IEEE Trans Pattern Anal Mach Intell*, vol. 41, no. 7, pp. 1720–1733, 2019, doi: 10.1109/TPAMI.2018.2845370.

[6] E. F. Abdelhafid, Y. M. Abdelkader, M. Ahmed, D. Rachid, and E. I. Abdelilah, "Visual and light detection and ranging-based simultaneous localization and mapping for self-driving cars," *International Journal of Electrical and Computer Engineering (IJECE)*, vol. 12, no. 6, p. 6284, 2022, doi: 10.11591/ijece.v12i6.pp6284-6292.

[7] J. Ni, K. Shen, Y. Chen, W. Cao, and S. Yang, "An improved deep network-based scene classification method for self-driving cars," *IEEE Trans Instrum Meas*, vol. 71, pp. 1–14, 2022, doi: 10.1109/tim.2022.3146923.

[8] Q. Zou, H. Jiang, Q. Dai, Y. Yue, L. Chen, and Q. Wang, "Robust Lane detection from continuous driving scenes using deep neural networks," *IEEE Trans Veh Technol*, vol. 69, no. 1, pp. 41–54, 2020, doi: 10.1109/tvt.2019.2949603.

[9] R. D. Brehar, M. P. Muresan, T. Marita, C.-C. Vancea, M. Negru, and S. Nedevschi, "Pedestrian street-cross action recognition in monocular far infrared sequences," *IEEE Access*, vol. 9, pp. 74302–74324, 2021, doi: 10.1109/access.2021.3080822.

[10] T. Alhadidi, A. Jaber, S. Jaradat, H. I. Ashqar, and M. Elhenawy, "Object Detection using Oriented Window Learning Vi-sion Transformer: Roadway Assets Recognition," Jun. 2024.

[11] L. Masello, G. Castignani, B. Sheehan, F. Murphy, and K. McDonnell, "On the road safety benefits of advanced driver assistance systems in different driving contexts," *Transp Res Interdiscip Perspect*, vol. 15, p. 100670, 2022, doi: 10.1016/j.trip.2022.100670.

[12] S. Niu, Y. Ma, and C. Wei, "Night-time lane positioning based on camera and LiDAR fusion," in *Sixth International Conference on Traffic Engineering and Transportation System (ICTETS 2022)*, SPIE, 2023, pp. 362–367.

[13] J. Orlovska, F. Novakazi, C. Wickman, and R. Söderberg, "Mixed-method design for user behavior evaluation of automated driver assistance systems: an automotive industry case," in *Proceedings of the Design Society International Conference on Engineering Design*, 2019, pp. 1803–1812. doi: 10.1017/dsi.2019.186.

[14] F. Abu Hamad, R. Hasiba, D. Shahwan, and H. I. Ashqar, "Driver Behavior at Roundabouts in Mixed Traffic: A Case Study Using Machine Learning," *J Transp Eng A Syst*, vol. 150, no. 12, p. 05024004, 2024.

[15] N. O. Khanfar, M. Elhenawy, H. I. Ashqar, Q. Hussain, and W. K. M. Alhajyaseen, "Driving behavior classification at signalized intersections using vehicle kinematics: Application of unsupervised machine learning," *Int J Inj Contr Saf Promot*, pp. 1–11, 2022.

[16] R. Pradhan, S. K. Majhi, J. K. Pradhan, and B. B. Pati, "Performance Evaluation of PID Controller for an Automobile Cruise Control System using Ant Lion Optimizer," *Engineering Journal*, vol. 21, no. 5, pp. 347–361, 2017, doi: 10.4186/ej.2017.21.5.347.

[17] D. Palac, I. Scully, R. Jonas, J. Campbell, D. Young, and D. Cades, "Advanced driver assistance systems (ADAS): who's driving what and what's driving use?," in *Proceedings of the Human Factors and Ergonomics Society Annual Meeting*, 2021, pp. 1220–1224. doi: 10.1177/1071181321651234.

[18] X. Huang, P. He, A. Rangarajan, and S. Ranka, "Intelligent intersection: Two-stream convolutional networks for real-time near-accident detection in traffic video," *ACM Transactions on Spatial Algorithms and Systems (TSAS)*, vol. 6, no. 2, pp. 1–28, 2020.

[19] H. I. Ashqar, T. I. AlHadidi, M. M. Elhenawy, and S. Jaradat, "Factors affecting crash severity in Roundabouts: A comprehensive analysis in the Jordanian context," *Transportation Engineering*, vol. 17, 2024, doi: 10.1016/j.treng.2024.100261.

[20] S. Jaradat, R. Nayak, A. Paz, H. I. Ashqar, and M. Elhenawy, "Multitask Learning for Crash Analysis: A Fine-Tuned LLM Framework Using Twitter Data," *Smart Cities*, vol. 7, no. 5, pp. 2422–2465, 2024, doi: 10.3390/smartcities7050095.

[21] S. Yang, B. Zhai, Q. You, J. Yuan, H. Yang, and C. Xu, "Law of Vision Representation in MLLMs," *arXiv preprint arXiv:2408.16357*, 2024.

[22] H. Sha et al., "Languagempc: Large language models as decision makers for autonomous driving," *arXiv preprint arXiv:2310.03026*, 2023.

[23] Z. Xu et al., "Drivegpt4: Interpretable end-to-end autonomous driving via large language model," *arXiv preprint arXiv:2310.01412*, 2023.

[24] H. I. Ashqar, A. Jaber, T. I. Alhadidi, and M. Elhenawy, "Advancing Object Detection in Transportation with Multimodal Large Language Models (MLLMs): A Comprehensive Review and Empirical Testing," *arXiv preprint arXiv:2409.18286*, 2024.

[25] Y. Ren, Y. Chen, S. Liu, B. Wang, H. Yu, and Z. Cui, "TPLLM: A traffic prediction framework based on pretrained large language models," *arXiv preprint arXiv:2403.02221*, 2024.

[26] O. Zheng, M. Abdel-Aty, D. Wang, Z. Wang, and S. Ding, "ChatGPT is on the horizon: could a large language model be suitable for intelligent traffic safety research and applications?," *arXiv preprint arXiv:2303.05382*, 2023.

[27] S. Masri, H. I. Ashqar, and M. Elhenawy, "Large Language Models (LLMs) as Traffic Control Systems at Urban Intersections: A New Paradigm," *arXiv preprint arXiv:2411.10869*, 2024.

## Appendix A

Please analyse the following image taken from an ego vehicle's perspective. For this image, answer each multi-choice question based on detected features, using '0' if a feature is not detected.

Multi-Choice Questions with Definitions:
1. Ambient: Refers to the time of day or lighting conditions in the image.
   - Values: Day (1), Dawn/Dusk (2), Night (3).
2. Attributes: Type of road attribute observed in the image.
   - Values: Straight road (1), Roundabout (2), Hilly road (3).
3. Construction Zone: Areas designated for roadwork, where special signs and barriers might be present.
   - Stages: Approaching (1), Entering (2), Passing (3).
4. Crosswalk: A designated pedestrian crossing area on the road, often marked with specific patterns, requiring vehicles to yield to pedestrians.

- Stages: Approaching (1), Entering (2), Passing (3).
5. Driveway: An entry or exit path to private property, such as a house or building, connecting to the main road.
   - Stages: Approaching (1), Entering (2), Passing (3).
6. Intersection (3-way): A point where three roads converge, requiring vehicles to manage merging, yielding, or stopping.
   - Stages: Approaching (1), Entering (2), Passing (3).
7. Intersection (4-way): A point where four roads converge, requiring vehicles to manage merging, yielding, or stopping.
   - Stages: Approaching (1), Entering (2), Passing (3).
8. Intersection (5-way or more): A point where five or more roads converge, requiring complex navigation or yielding.
   - Stages: Approaching (1), Entering (2), Passing (3).
9. Overhead Bridge/Underpass: Structures that allow one road to pass over or under another.
   - Stages: Approaching (1), Entering (2), Passing (3).
10. Tunnel: A covered section of road, often through a hill or mountain.
    - Stages: Approaching (1), Entering (2), Passing (3).
11. Rail Crossing: A point where a railway track intersects with the road.
    - Stages: Approaching (1), Entering (2), Passing (3).
12. Surface: Condition of the road, which can impact driving safety.
    - Values: Dry (1), Wet (2), Icy (3), Snow (4).
13. Types: Classification of the road type based on surroundings and purpose.
    - Values: Local (1), Highway (2), Ramp (3), Urban (4), Rural (5).
14. Weather: Observed weather conditions affecting visibility and road conditions.
    - Values: Sunny (1), Cloudy (2), Rain (3), Snow (4), Fog (5), Hail (6).
15. No-Signal Intersection: An intersection without traffic lights, where vehicles yield based on rules.
    - Stages: Approaching (1), Passing (2).
16. Stop Intersection: An intersection with stop signs, requiring a full stop by vehicles.
    - Stages: Approaching (1), Passing (2).
17. Merge/Gore on Left: A triangular section at the convergence or divergence of roadways on the left, where lanes merge or split.
    - Stages: Approaching (1), Passing (2).
18. Merge/Gore on Right: A triangular section at the convergence or divergence of roadways on the right, where lanes merge or split.
    - Stages: Approaching (1), Passing (2).
19. Branch/Gore on Left: A split in the road where traffic divides to the left, often marked with signage or painted lines.
    - Stages: Approaching (1), Passing (2).
20. Branch/Gore on Right: A split in the road where traffic divides to the right, often marked with signage or painted lines.
    - Stages: Approaching (1), Passing (2).
21. Zebra Crossing: A type of pedestrian crossing marked by white stripes, where vehicles must yield to pedestrians.
    - Stages: Approaching (1), Passing (2).

Instructions: For this image, provide the analysis in JSON format as shown below, assigning each category a value of '0' if the feature is not detected:

```
{
  "Ambient": 0,
  "Attributes": 0,
  "Construction_zone": 0,
  "Cross_walk": 0,
  "Driveway": 0,
  "Intersection (3 way)": 0,
  "Intersection (4 way)": 0,
  "Intersection (5 way & more)": 0,
  "Overhead_bridge/under_overpass": 0,
  "Tunnel": 0,
  "Rail_crossing": 0,
  "Surface": 0,
  "Types": 0,
  "Weather": 0,
  "NoSignalIntersection": 0,
  "StopIntersection": 0,
  "Merge_GoreOnLeft": 0,
  "Merge_GoreOnRight": 0,
  "Branch_GoreOnLeft": 0,
  "Branch_GoreOnRight": 0,
  "ZebraCrossing": 0
}
```